\documentclass{article}
\usepackage{amssymb}
\usepackage{wrapfig}
\usepackage{pifont}
\usepackage{graphicx}
\usepackage{caption}
\usepackage{subcaption}
\usepackage{xcolor}
\usepackage{booktabs}
\usepackage{multirow}

\newcommand{\cmark}{\textcolor{green!60!black}{\ding{51}}}  
\newcommand{\xmark}{\textcolor{red}{\ding{55}}}             
\usepackage[T1]{fontenc}
\usepackage[utf8]{inputenc}
\usepackage{hyperref}
\usepackage{color}
\usepackage{verbatim}
\usepackage{float}
\usepackage{amsmath}

\usepackage{amsthm}
\usepackage{amssymb}
\usepackage{stmaryrd}
\usepackage{stackrel}
\usepackage{graphicx}
\usepackage{wasysym}
\usepackage{mathtools}
\usepackage{hyperref}
\usepackage{url}   
\usepackage{booktabs} 
\usepackage{nicefrac} 
\usepackage{multirow}
\usepackage{xcolor}
\usepackage{microtype}
\usepackage{paralist}
\usepackage{subcaption}
\makeatletter
\usepackage{float}

\floatstyle{ruled}
\newfloat{algorithm}{tbp}{loa}
\providecommand{\algorithmname}{Algorithm}
\floatname{algorithm}{\protect\algorithmname}

\theoremstyle{plain}

\theoremstyle{definition}

\theoremstyle{definition}
\newtheorem*{problem}{\protect\problemname}
\theoremstyle{plain}

\theoremstyle{definition}
\newtheorem{example}{\protect\examplename}
\theoremstyle{remark}

\theoremstyle{plain}

\usepackage{cite}
\usepackage{algpseudocode}
\usepackage{setspace}

\makeatother

\usepackage{babel}
\providecommand{\corollaryname}{Corollary}
\providecommand{\definitionname}{Definition}
\providecommand{\examplename}{Example}
\providecommand{\lemmaname}{Lemma}
\providecommand{\problemname}{Problem}
\providecommand{\remarkname}{Remark}
\providecommand{\theoremname}{Theorem}

\newcommand{\E}{\mathbb{E}}
\newcommand{\setS}{\mathcal{S}}
\newcommand{\setA}{\mathcal{A}}
\newcommand{\setP}{\mathcal{P}}

\newcommand{\setX}{\mathcal{X}}

\newcommand{\setR}{\mathbb{R}}

\newcommand{\V}{V}

\newcommand{\x}{\mathbf{x^*}}
\newtheorem{theorem}{Theorem}

\newtheorem{proposition}{Proposition}

\usepackage[preprint]{corl_2025} 

\title{Bridging Deep Reinforcement Learning and \\Motion Planning for Model-Free \\ Navigation in Cluttered Environments}

\author{
  \makebox[0.45\textwidth][c]{Licheng Luo} \hfill
  \makebox[0.45\textwidth][c]{Mingyu Cai} \\
  \makebox[0.45\textwidth][c]{Department of Mechanical Engineering} \hfill
  \makebox[0.45\textwidth][c]{Department of Mechanical Engineering} \\
  \makebox[0.45\textwidth][c]{University of California Riverside} \hfill
  \makebox[0.45\textwidth][c]{University of California Riverside} \\
  \makebox[0.45\textwidth][c]{\texttt{lichengl@ucr.edu}} \hfill
  \makebox[0.45\textwidth][c]{\texttt{mingyuc@ucr.edu}}
}

\begin{document}
\maketitle


\begin{abstract}
    Deep Reinforcement Learning (DRL) has emerged as a powerful model-free paradigm for learning optimal policies. However, in navigation tasks with cluttered environments, DRL methods often suffer from insufficient exploration, especially under sparse rewards or complex dynamics with system disturbances.  To address this challenge, we bridge general graph-based motion planning and DRL, enabling agents to explore cluttered spaces more effectively and achieve desired navigation performance. Specifically, we design a dense reward function grounded in a graph structure that spans the entire state space. This graph provides rich guidance, steering the agent toward optimal strategies. We validate our approach in challenging environments, demonstrating substantial improvements in exploration efficiency and task success rates.  
    The project website is available at: \href{https://plen1lune.github.io/overcome_exploration/}{https://plen1lune.github.io/overcome\_exploration/}
\end{abstract}

\keywords{Deep Reinforcement Learning, Motion Planning, Model-free Control} 


\section{Introduction}

Deep Reinforcement Learning (DRL) provides a flexible and general approach for acquiring control policies in tasks characterized by high-dimensional observations and complex, uncertain dynamics~\citep{arulkumaran2017brief,sutton2018reinforcement}. Its \emph{model-free} nature enables agents to learn directly from interactions, eliminating the need for explicit dynamics modeling. This makes DRL particularly suitable for domains such as robotic manipulation~\citep{levine2016end,kalashnikov2018qt}, autonomous driving~\citep{kendall2019learning}, and locomotion~\citep{heess2017emergence}, where accurate models can be difficult to obtain. In recent studies, DRL demonstrates strong potential for generalization and robustness, supporting intelligent decision-making under diverse and dynamic conditions~\citep{yao2024sonic}.

However, the practical success of DRL remains severely limited by its sample inefficiency, primarily due to inadequate exploration in sparse-reward or high-dimensional settings~\citep{ecoffet2021first,bellemare2016unifying,zhang2025lamma}. Without informative feedback, agents may spend substantial time interacting with irrelevant states. Numerous strategies have been proposed to improve exploration in DRL, including intrinsic rewards~\citep{pathak2017curiosity,burda2019exploration}, density-based bonuses~\citep{tang2017exploration,bellemare2016unifying}, entropy regularization~\citep{haarnoja2018soft}, policy perturbation~\citep{schulman2017proximal}, and hierarchical or logic-guided approaches~\citep{2021caiModular,kantaros2024sampleefficientreinforcementlearningtemporal}. These approaches have partially alleviated exploration challenges and improved sample efficiency in regular environments. However, these methods still struggle with sparse reward signals and high-dimensional, cluttered environments—challenges that remain largely unsolved.

The primary contribution of this work is a graph-based exploration framework that enables efficient navigation from arbitrary initial states to task goals in cluttered environments. We address three key challenges that prior methods leave unresolved: 1) Bridging the gap between motion-planning feedback and the Markovian reward structures required by model-free RL algorithms; 2) handling infeasible planning paths and tracking errors; 3) enabling generalization to unseen cases.
Our method builds on a classical motion planning backbone to provide informative guidance for exploration. It is lightweight, compatible with a range of model-free RL algorithms, and requires no policy modification across different initializations. We further provide theoretical justification that our exploration strategy preserves the original RL objective and empirically show improved sample efficiency and task success across a range of cluttered environments.

\section{Related Work}
\label{sec:relatedworks}
\noindent
\textbf{Motion Planning:}  
Sampling-based planners such as RRT, RRT*, PRM, and RRG~\citep{lavalle2001randomized,karaman2011sampling,kavraki1996probabilistic} are widely used for collision-free path planning in continuous spaces. PRM builds a reusable roadmap for multi-query tasks by connecting sampled points with feasible edges, while RRG extends RRT with a graph structure for asymptotic optimality. Graph search algorithms like A*~\citep{hart1968formal} are typically used to compute cost-optimal paths on these graphs. Although effective, these planners usually require model-based controllers to execute the planned trajectories.

\vspace{0.1cm}
\noindent
\textbf{Deep Reinforcement Learning:}
DRL has emerged as a powerful model-free paradigm for learning optimal policies in complex decision-making problems~\citep{sutton2018reinforcement}. In continuous control tasks, DRL algorithms such as Proximal Policy Optimization (PPO)~\citep{schulman2017proximal} and Soft Actor-Critic (SAC)~\citep{haarnoja2018soft} have been widely adopted due to their empirical stability and sample efficiency. These algorithms are typically built upon the actor-critic framework, which has become the standard paradigm in modern policy-based reinforcement learning~\citep{sutton2018reinforcement,haarnoja2018soft}. In this framework, the \textit{actor} is a parameterized policy that maps states to actions, while the \textit{critic} commonly estimates the expected return—either in the form of a state-value function or an action-value function—to guide the improvement of the actor. 

\vspace{0.1cm}
\noindent
\textbf{Exploration Strategies:}  
A broad class of methods augment the reward signal to improve exploration. Curiosity-driven methods use forward dynamics models to measure prediction error as an intrinsic reward signal for novelty~\citep{pathak2017curiosity,burda2019exploration,schwarke2023curiosity}, encouraging agents to seek transitions that are poorly understood. Count-based exploration estimates state novelty through density modeling~\citep{tang2017exploration,bellemare2016unifying}, which generalizes tabular counts to continuous domains. Entropy-based methods~\citep{haarnoja2018soft} promote stochastic policies by maximizing uncertainty in action selection, while methods like PPO~\citep{schulman2017proximal} introduce noise via clipped policy updates. Modular methods~\citep{2021caiModular} introduce abstract goals or subtasks to guide long-horizon behavior, and diffusion-based methods~\citep{wang2023cold} generate trajectories from good replay states to guide planning-based exploration.
\citet{balsells2023autonomous} introduce the notion of reachability, whereas \citet{gieselmann2023expansive} consider the latent space in their formulation. More recently, formal methods such as temporal logic have been used to encode high-level constraints~\citep{kantaros2024sampleefficientreinforcementlearningtemporal}. While these methods have shown empirical success in well-structured or open environments, they often struggle in cluttered settings where disconnected free spaces, narrow passages, or bottlenecks make it difficult for agents to reach informative states. 
The most closely related work is~\citep{cai2022overcoming}, which leverages RRT-based guidance to improve exploration. However, their method is fundamentally constrained by the feasibility of constructing valid planning trajectories, and lacks the extensibility needed to handle unseen obstacles or dynamic changes, both of which are addressed in our work.

\section{Preliminary and Problem Formulation}
\label{sec:Problem}
\vspace{0.1cm}
\noindent
\textbf{System Disturbance:}
We consider a continuous-time dynamical system $S$ with general unknown dynamics, whose evolution is described by
$\frac{ds}{dt} = \dot{s} = f(s,a,d)$
where $s\in\setS\subseteq\setR^n$ represents the state variables, $a\in\setA\subseteq\setR^m$ denotes the control inputs, and $d$ is the disturbance function.

\vspace{0.1cm}
\noindent
\textbf{MDPs:} Markov Decision Processes (MDPs) provide a mathematical framework for modeling sequential decision-making problems in stochastic environments. In the discounted setting, we consider a tuple 
$(\setS,\setA,p,r,\rho_0,\gamma)$, 
where $\setS$ denotes a continuous state space, 
$\setA$ denotes an continuous action space, 
$p:\setS\times\setS\times\setA\to[0,1]$ denotes a transition function, 
\( r: \mathcal{S} \times \mathcal{A} \to [r_{\min}, r_{\max}] \) be a bounded reward function.
$\rho_0:\setS\to[0,1]$ denotes an initial distribution, 
and $\gamma\in(0,1]$ is a discount factor.
At the beginning of the process, 
the agent's initial state $s_0$ is sampled 
from the initial state distribution $\rho_0$. 
At each time step $t$, the agent observes the current state $s_t$, 
and selects an action $a_t$ based on its policy $\pi:\setS\to\setP(\setA)$, 
which is a mapping from states to probability distributions on $\setA$. 
The agent samples an action according to the probability distribution, 
i.e., $a_t \sim \pi(\cdot | s_t)$. 
Once the agent takes the action $a_t$, 
the environment responds by transitioning the agent 
to a new state $s_{t+1}$, 
governed by the transition function $p$, 
i.e., $s_{t+1} \sim p(\cdot | s_t, a_t)$. 
After that, it receives a scalar reward $r_{t+1}$ from the environment. 
This reward reflects the immediate benefit or cost 
of taking action $a_t$ in state $s_t$. 
To balance immediate and future rewards, 
the agent uses the discount factor $\gamma$, 
which ensures that rewards received earlier in time 
are more valuable than those in the distant future. 
The total discount reward, also called the return, is calculated as
$G_t = \sum_{k=0}^{\infty} \gamma^k r(s_{t+k}, a_{t+k})$,
where $G_t$ denotes the return at the time step $t$.
Given the start state $s$ and policy $\pi$, 
we define the state value function as
$V_{\pi}(s) = \E_{\pi,p}\left[G_t | s_t = s\right]$.

\begin{figure}[H]
\begin{center}
\centerline{\includegraphics[width=\columnwidth]{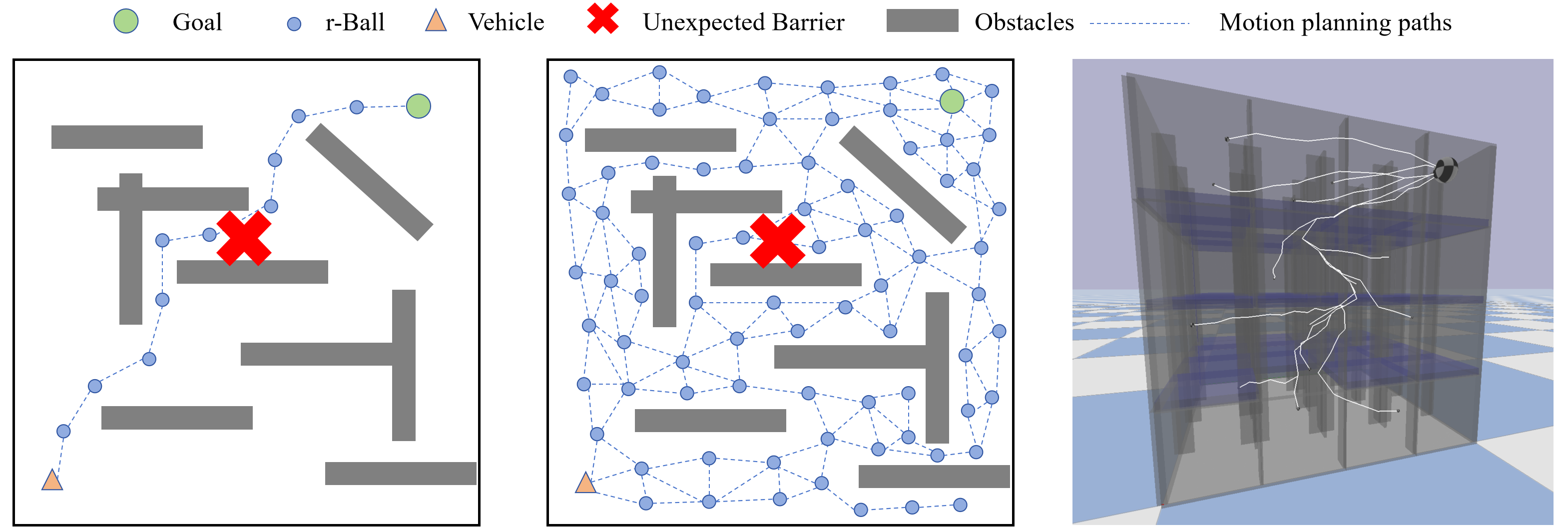}}
\caption{Comparison between previous method and graph-based method. We implemented both RRT (left) and RRG (middle) to solve the same goal-reaching task. As shown, the gray areas represent known obstacles, and the red cross indicates potential disturbances that may lead to failure transitions.
The right figure shows a 3D navigation environment where graph-based planning generates feasible paths (white lines) through complex obstacle configurations. The translucent walls represent static barriers in the 3D space, demonstrating the scalability of our approach to high-dimensional planning problems.}
\label{fig:comparison}
\vspace{-2em}
\end{center}
\end{figure}

\begin{problem}
\label{problem1}
We focus on finding a feasible solution in challenging environments. While standard MDP theory guarantees the existence of an optimal policy under mild conditions, learning such a policy via model-free RL remains difficult in sparse-reward or long-horizon settings. These challenges are further exacerbated in practice by the clutter of the environment. Our work specifically addresses these issues by enhancing exploration through a graph-based guidance framework.
\end{problem}
\vspace{-0.5em}

\begin{example}
We compared our approach with methods that use motion planning-generated paths to guide the training of reinforcement learning agents~\citep{cai2022overcoming} (see Figure~\ref{fig:comparison}). To illustrate their limitations, we deliberately introduced an infeasibility (indicated by a red cross) along the planned geometric path to the goal. The tree-based method (left) fails to manage such disturbances, as it relies on a single, fixed trajectory without considering alternative routes, resulting in failure. Consequently, using waypoints from this path as guidance also fails. In contrast, our approach employs a graph that covers the entire workspace (middle), providing more robust guidance.
\end{example}

\section{Overcoming Exploration\label{sec:exploration_Solution}
}

In Section~\ref{sec:exploration_Solution}, we introduce the graph-based reward and our design enabling arbitrary starting points.

\subsection{Geometric Graph-based Motion Planning Methods}
While our implementation employs RRG for demonstration purposes, the proposed framework is fundamentally agnostic to the choice of graph construction methods. In particular, it accommodates both sampling-based planners, such as RRG~\citep{karaman2011sampling} and PRM~\citep{kavraki1996probabilistic}, and heuristic-guided search algorithms, notably A*~\citep{hart1968formal}.
The RRG incrementally constructs an undirected graph $G = (V, E)$ in the configuration space $\mathcal{X}$, where $V$ is the set of vertices representing sampled points, and $E$ consists of edges that connect pairs of vertices via collision-free paths. Compared to tree-based methods, such as RRT, RRG maintains cycles and improves connectivity by linking new samples to their nearest neighbors, thereby providing a denser approximation of the configuration space. This graph structure serves as the foundation for our method, and we briefly introduce two key functions that will be used in the following sections.

\textbf{Steering Function:} Given two states $ s $ and $ s' $, the function $\text{Steer}(s, s', \eta)$ returns a new state $ s_{\text{new}} $ such that $ s_{\text{new}} $ lies on the path connecting $ s $ to $ s' $ and the distance $ \|s- s_{\text{new}}\|_2 \leq \eta $. The step size $ \eta $ is a user-defined parameter that controls how far the path extends in a single step.

\textbf{Sampling and Connection:} The sampling function generates a new state $ s_{\text{new}} $ in the free space, and the connection function attempts to connect $ s_{\text{new}} $ to its nearest neighbors $ \mathcal{N}(s_{\text{new}}) $ using a geometric trajectory. For RRG, if a connection is successful, the edge is added to the graph, forming a cycle when possible.

Given the graph $G = (V, E)$ constructed by RRG, if a node $x \in V$ lies within the goal region $\mathcal{S}_G$, we compute the optimal state trajectory $\mathbf{x}^*$ as a sequence of geometric states:
\(
\mathbf{x}^* = \{x_0, x_1, \ldots, x_{d}\},
\)
where $x_i\in\V\subset\setS$ denotes nodes and $x_{d} \in \mathcal{S}_G$. In most planning pipelines, Dijkstra's algorithm~\citep{dijkstra1959note} is employed to extract the shortest waypoint sequence from the constructed graph.
Beyond sampling-based graphs, grid-based graphs constitute another practical instantiation of our framework. In such cases, connectivity is predefined, and shortest paths are typically computed using algorithms such as A*. Additional details are provided in Appendix~\ref{append:cost2goal}.

\subsection{Motion Planning Guided Reward}

After employing a graph-based motion planning method, we obtained a set of waypoints that span the entire graph. Based on this, we use an optimal path algorithm to compute the distance from each node to the goal region. We denote the optimal trajectory by $\x$, and the optimal is defined according to the length of trajectories. $s|\x$ denotes a state of path $\x$. 
We define the distance function $Dist: \setS\times\setS\times\setX\rightarrow [0,\infty]$ to represent the Euclidean distance between two states along a given path, and define the cost function $Cost: \setS\times\setX\rightarrow[0,\infty]$ to represent the distance between one state and the destination $s_d$. Hence, the distance between two states along the optimal path can be expressed as $Dist(s_1,s_2|\x) = |Cost(s_1|\x)-Cost(s_2|\x)|$. Consequently, the graph $ G $ is now extended to $ G = (V, E, Cost) $, where we slightly abuse the notation, denoting $Cost$ as distance between the current node and the goal.

The distance along the path enables us to accurately quantify the actual path length, making it more applicable in cluttered environments compared to the Euclidean distance. Based on this distance, we design the motion-planning reward scheme to guide the RL agent. Given the difficulty of reaching an exact state in continuous space, we define the norm r-ball to loosen the criteria for determining whether the agent has reached specific states. Formally, it is defined by $Ball_r(s) = \left\{s'\in\setS\mid\|s-s'\|_2\leq r\right\}$, where $r$ is the radius. The selection of the r-ball radius can influence the performance of the algorithm. For simplicity, we use the default $r= \frac{\eta}{2}$ based on the steering function in all the following sections.
We then define a progression function $D:\setS\rightarrow [0,\infty]$ to determine whether the agent has moved toward the goal region and got closer to it. Formally,
\begin{equation}
    D(s) = 
    \begin{cases}
       Cost\{s_i|\x\}, & \text{if } x \in \text{Ball}_r(s_i \mid \x) \\
        \infty, & \text{otherwise}
    \end{cases}
\end{equation}

We adopt the definition of history from~\citep{puterman2014markov} with a slight modification $h_t = \{s_1,a_1,\dots,s_{t-1}, a_{t-1}\}$ instead of $\{s_1,a_1,\dots,s_{t-1}, a_{t-1},s_t\}$ , thereby the $D_{min}$ over a state-action sequence $\tau_t = \{h_t,s_t,a_t\}$ is defined as 
$
    D_{min}(\mathbf{\tau}) = \min_{s\in\mathbf{s}}{D(s)}
$, where $\mathbf{s}$ denotes the state sequence in trajectory $\tau_t$.
Ideally, we would check the change of $ D_{min} $ at each step to determine whether the agent has moved closer to the goal region, i.e., check if $D_{min}(\tau_t) < D_{min}(\tau_{t-1})$, and assign a positive reward $R_+$ accordingly. 
However, this design introduces a non-Markovian reward issue. In other words, $r(h_t,s_t,a_t)\neq r(h'_t,s_t,a_t)$ could happen with different history $h_t$ and $h'_t$. 
To address this problem, we denote $ \setS^{\times} = \{ (s, D_{min}) \mid s \in \setS, \, D_{min} \in \mathbb{R} \} $, so that $ D_{min} $ can be directly retrieved from the augmented state. 
Correspondingly, the \textbf{\textit{augmented reward function}} is defined as \( \tilde{r}: \setS^{\times} \times \setA \times \setS^{\times} \to \mathbb{R} \), 
\begin{equation}
\tilde{r}_t(s_{t}^{\times}, a_t, s_{t+1}^{\times}) = \left\{ 
\begin{array}{ll}
R_{-}, & \text{if } s_{t+1} \text{ is in the obstacles}, \\
R_{++}, & \text{if } s_{t+1} \text{ is in the goal area}, \\
R_{+}, & \text{if } D_{min}^{t+1} < D_{min}^{t}, \\
0, & \text{otherwise.}
\end{array}
\right.
\label{eq:reward_function}
\end{equation}
where $R_{++}\gg R_{+}$ prioritizes goal achievement as the primary objective, while $R_{+}$ incentivizes exploratory progress toward the goal.
We next analyze the properties of the reward function in this augmented state space.

\begin{theorem}\label{th:markovian}
\emph{(Markovian Analysis)} 
By extending the state space to $\setS^{\times}$, we ensure that the augmented reward function $\tilde{r}$ shown in Eq.~\eqref{eq:reward_function} satisfies the Markov property. Formally,
\[
   P\bigl(\tilde{r}_t \mid s_0^\times,a_0,\dots,s_t^\times,a_t,s_{t+1}^\times\bigr) 
   \;=\;
   P\bigl(\tilde{r}_t \mid s_t^\times,a_t,s_{t+1}^\times\bigr)
\]
for all time steps $t$, where $s_t^{\times} \in \setS^{\times}$ denotes the augmented state including $D_{min}$.
\end{theorem}

This result ensures that our planning-augmented reward remains compatible with standard RL algorithms. In the remainder of this paper, we operate in the augmented state space $\mathcal{S}^{\times}$ to maintain this property. The detailed proof is presented in Appendix~\ref{append:markovianproof}.

\begin{proposition}\label{prop:optimalinvariance}
\emph{(Reward Performance)}  
Optimizing the reward defined in Eq.~\eqref{eq:reward_function} accelerates convergence and yields consistently high success rates in cluttered environments.
\end{proposition}

This proposition guarantees that the auxiliary rewards accelerate convergence without altering the original task objective.
The proof is presented in Appendix~\ref{append:optimalinvarianceproof}. 
The learning objective aims to find the optimal policy by optimizing the parameters $\theta$: $$
\theta^* = \arg\max_{\theta} J(\theta) = \mathbb{E}_{\pi_\theta} \left[ \sum_{t=0}^{\infty} \gamma^t \cdot \tilde{r}_t(s_{t}^{\times}, a_t, s_{t+1}^{\times}) \right]
$$
which corresponds to minimizing the following loss function:
\[
\theta^* = \arg\min_{\theta} \mathbb{E}_{(s^\times, a, r, s'^\times) \sim \mathcal{D}} \left[ \left(Q_{\omega}(s, a) - y \right)^2 \right]
\]
where $y$ is the target value, $\mathcal{D}$ is the replay buffer and $Q_{w}$ denotes the action-value function parameterized by $\omega$. It is worth noting that almost all reinforcement learning algorithms designed for cluttered environments adopt the actor-critic architecture~\citep{lillicrap2015continuous, haarnoja2018soft, fujimoto2018addressing, schulman2017proximal}.
As noted in~\citep{lillicrap2015continuous}, actor-critic methods are sensitive to the quality and distribution of samples stored in the replay buffer, as effective optimization of the loss function depends heavily on informative and diverse experiences. Our proposed motion-planning based reward exhibits high spatial density, which facilitates exploration even under stochastic or suboptimal policies. Consequently, this promotes more stable and sample-efficient learning throughout training.

\subsection{Infeasible Paths Analysis}

Recalling Sec.~\ref{sec:Problem}, our method specifically addresses the challenge of unexpected disturbances in the simulation, which often render paths infeasible and cause tracking errors while training. In contrast to tree-based methods, which focus on planning a single path without accounting for its robustness or feasibility, our approach leverages a graph-based motion planning strategy. 

We highlight that the robustness of our method stems from the redundancy introduced by graph-based planning. Rather than following a single predetermined trajectory, the agent can flexibly choose among multiple feasible alternatives toward the goal. This redundancy is particularly beneficial in the presence of disturbances or partial map mismatches, where some paths may become infeasible during execution. As the graph becomes denser—either through increased sampling resolution or stronger connectivity—the number of distinct paths between the start and goal increases. Prior work in random geometric graphs shows that the connectivity and availability of alternative paths grow rapidly with the number of nodes and connection radius~\citep{penrose2003random}. This property greatly enhances the likelihood that at least one safe and feasible path exists, improving resilience against both local failures and global perturbations.\\
\begin{wrapfigure}{r}{0.5\textwidth}
\centering
\begin{minipage}{0.24\textwidth}
    \centering
    \includegraphics[width=\linewidth]{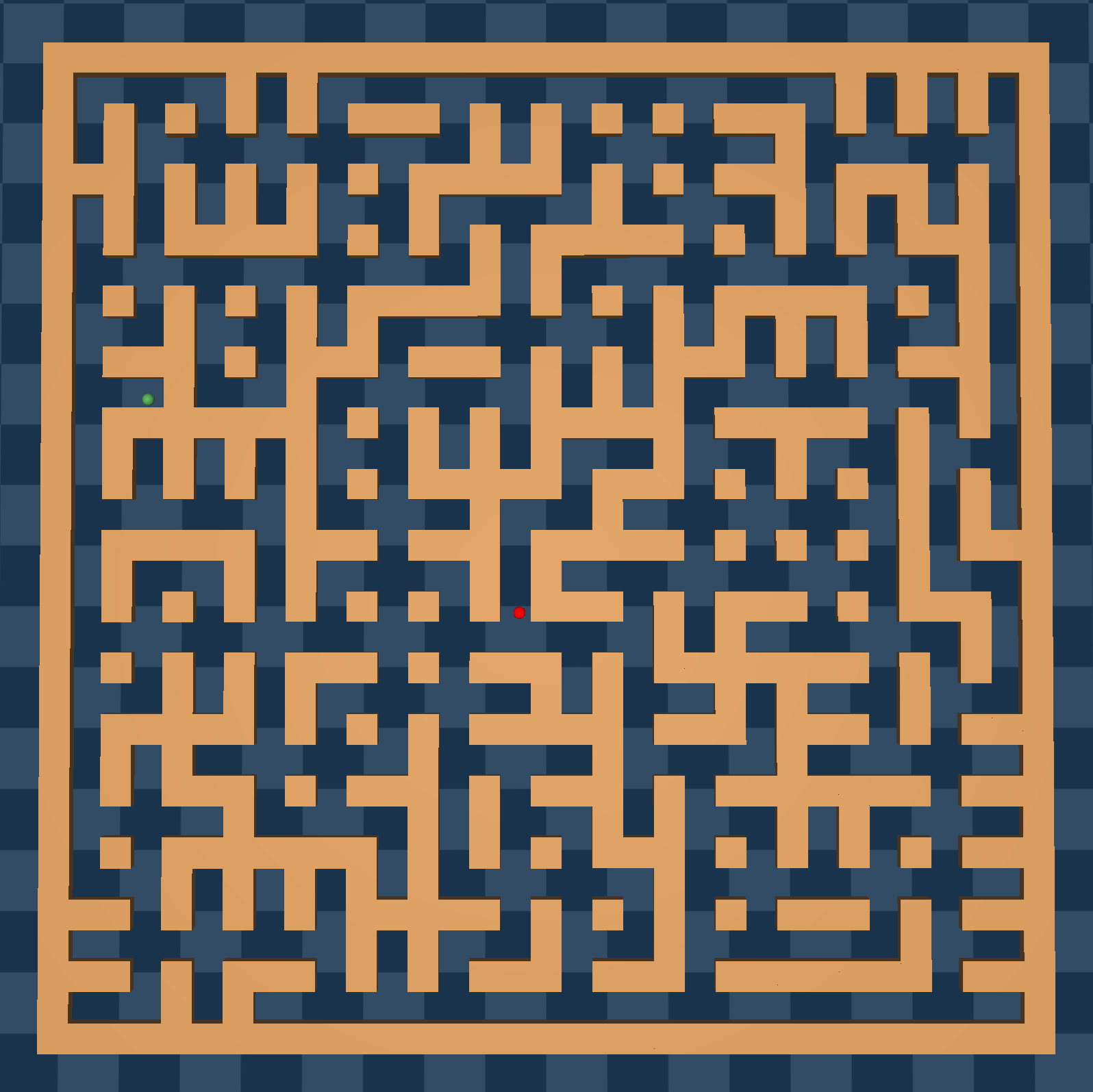}
\end{minipage}
\hfill
\begin{minipage}{0.24\textwidth}
    \centering
    \includegraphics[width=\linewidth]{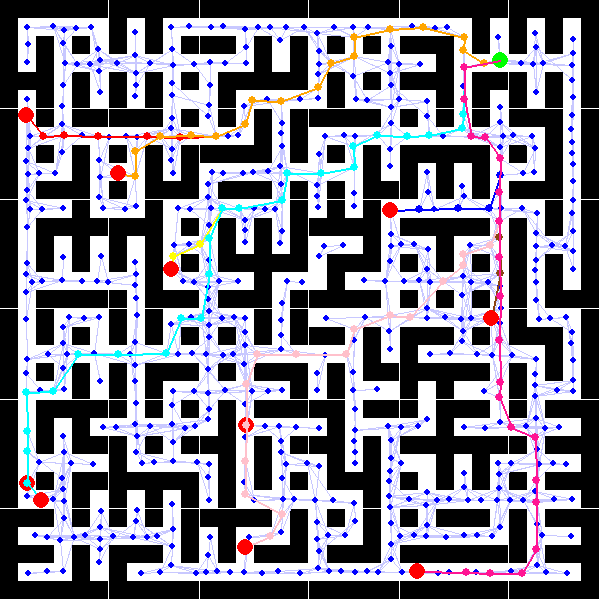}
\end{minipage}
\caption{Left: Maze layout. Right: Diverse goal-reaching paths from nearby starting points.}
\vspace{-2em}
\label{fig:maze}
\end{wrapfigure}
\begin{example}
    As illustrated in Figure~\ref{fig:maze} (right), in a complex but highly connected maze, our graph-based method can easily identify alternative paths to the goal. Even from nearby but distinct start positions, the agent successfully reaches the same goal via different trajectories, demonstrating the diversity and flexibility of path selection enabled by dense connectivity.
\end{example}
\subsection{Advantages and Extensions}\label{sec:generalize}
With the graph-based waypoints covering the whole environment, we could naturally extend the training objective from single start point to arbitrary start points. This brings some desirable properties that enable multiple extensions.

\textbf{Scalable Control:}
In an autonomous system with tasks described by temporal logic formulas, if there are $ N $ goal regions, in the worst case, if a neural network controller can only move from a specific starting point to a specific endpoint, the system would require $ O(N^2) $ distinct controllers. This indicates that in complex systems, the demand for controllers will increase dramatically as the number of destinations grows. A similar limitation is discussed in \citep{cai2022overcoming}, where a separate controller is trained for each unique start-goal pair.
However, by generalizing the initialization, the required number of controllers is only $ O(N) $, i.e., only one controller is needed for each goal region. This generalization is particularly beneficial for LTL-specified tasks, a representative setting in cluttered environments, as it substantially mitigates onboard resource limitations and enhances system scalability.

\textbf{Handling Unseen Situations:}
The ability to reach goals from arbitrary initializations can be further strengthened by integrating recovery mechanisms, such as safety filters—preferably model-free ones—as well as fallback policies or local replanning modules, to handle unexpected disturbances and transform otherwise infeasible scenarios into solvable tasks. This capability provides a significant advantage over prior methods, inherently enhancing system robustness and supporting generalization to new scenarios. Although recovery is not the main focus of this work, we include a simple distance-based filter in our experiments to illustrate the concept.

\section{Experimental Results\label{sec:experiment}}
We evaluated both our method and the baseline approaches using widely adopted reinforcement learning algorithms, i.e., PPO. We use a high-quality implementation of \citet{stable-baselines3}, with most hyperparameters left at their default settings.
The results show that our approach significantly improves the success rate and training efficiency.

\begin{figure}[htbp]
  \centering
  \begin{minipage}{\textwidth}
    \centering
    \includegraphics[width=\linewidth]{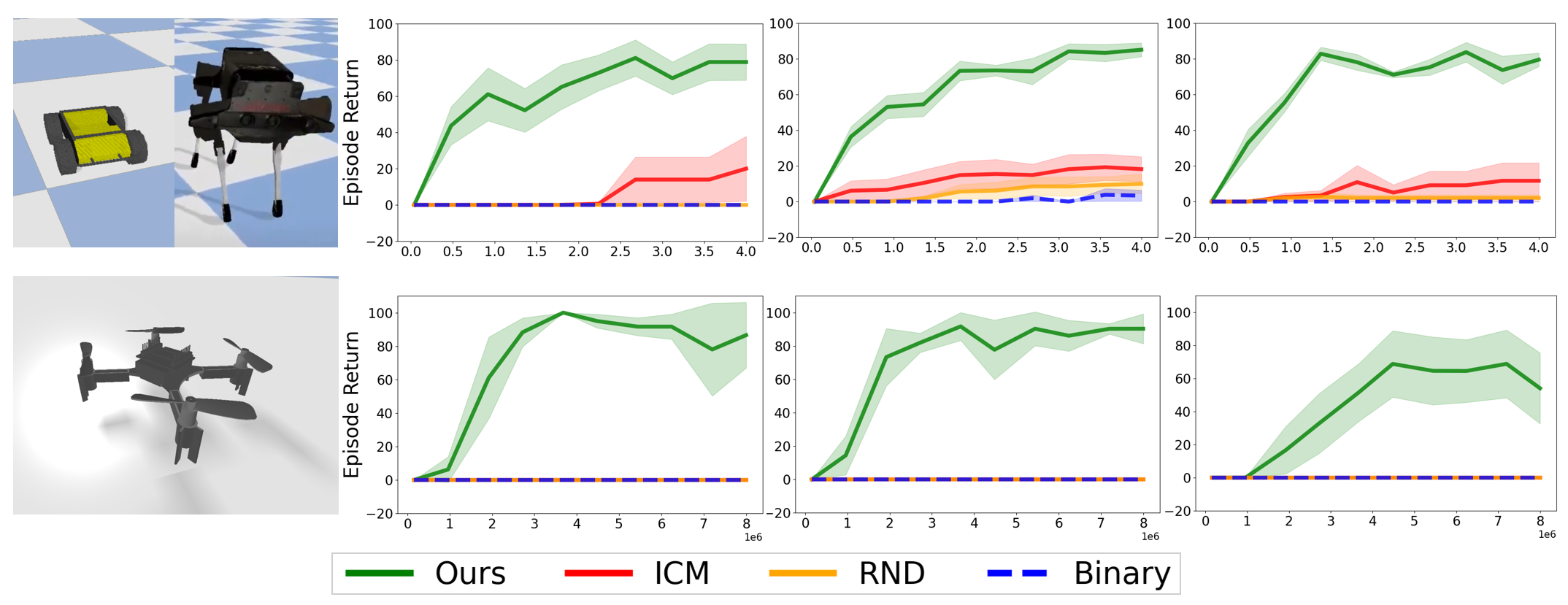}
  \end{minipage}
  \caption{Comparison of training efficiency (consumption) across different methods. Each plot shows the mean and standard error over 10 runs. Methods that do not appear in certain parts of the curves indicate overlap at the bottom.}
  \vspace{-2em}
  \label{fig:consumption_comparison}
\end{figure}

\textbf{Baseline Approaches:}
We conducted experiments in both 2D and 3D cluttered environments across different scales. To benchmark our approach, we compared against representative methods from count-based and curiosity-driven exploration, namely ICM~\citep{pathak2017curiosity} and RND~\citep{burda2019exploration} in cluttered environments. Additionally, we evaluated our method against the RRT-guided reward~\citep{cai2022overcoming} and a distance-based reward (referred to as binary) to assess its robustness in navigating under disturbances.

\textbf{Dynamic Model:}
We evaluate our method across two representative dynamic models implemented using the PyBullet physics engine~\citep{coumans2016pybullet}, corresponding to distinct control modes. The first is a quadrotor model controlled via velocity commands in 3D space (i.e., $[v_x, v_y, v_z]$), while the second includes a car-like ground vehicle and a quadruped robot, both controlled using high-level commands in planar space (i.e., $[v_x, \dot{\psi}]$).
While accurate modeling and low-level control of these systems are not the focus of this work, such components can be readily obtained from prior work on robust locomotion and dynamics learning~\citep{rudin2022learningwalk, panerati2021learning}, or replaced with task-specific pretrained controllers depending on the target platform.
\begin{wrapfigure}{r}{0.5\textwidth}
    \centering
    \includegraphics[width=\linewidth]{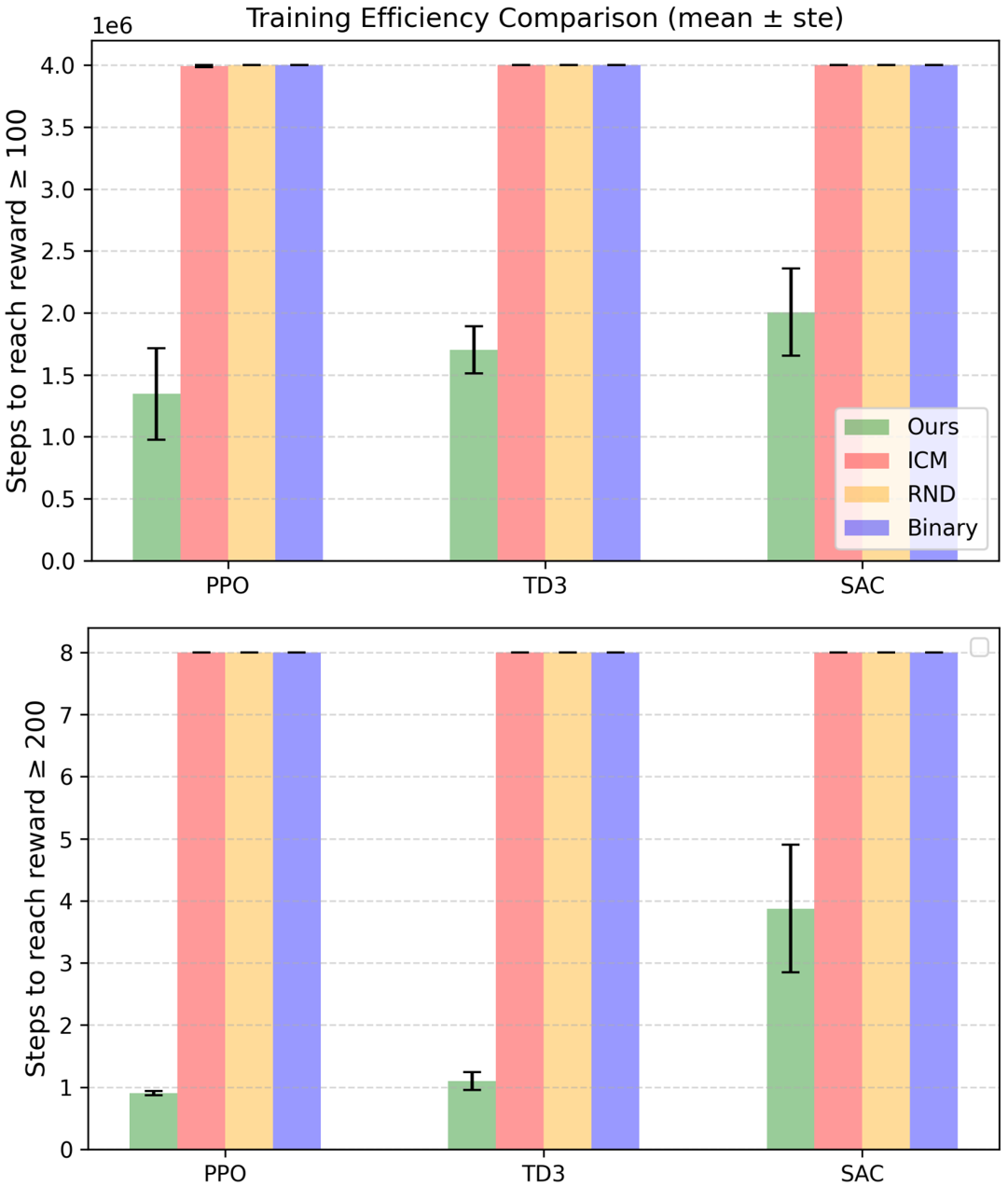}
    \caption{Comparison with other reward functions. The bar height (lower is better) indicates the number of steps required for the agent to complete the task for the first time during training.}
    \vspace{-1em}  
    \label{fig:efficiency}
\end{wrapfigure}
\textbf{Task Design:}  
We consider a goal-reaching task where the agent must navigate from various initial positions to a designated goal region.  
Both the 2D and 3D environments are structured as mazes, where vanilla distance-based rewards are insufficient for effective navigation.  
Further, the 3D environment is constructed as a $4\times4\times4$ room grid, where each floor contains only a single passage leading to the next level, and rooms within the same floor are interconnected by doors.  
The goal is placed at the topmost layer, while starting positions are uniformly sampled across the lower three floors. 
Additionally, for comparisons with the RRT-based method~\citep{cai2022overcoming}, we randomly block three doors on the same floor to simulate scenarios where path planning becomes infeasible.

\textbf{Efficient Exploration:}  
We consider each training task as a compositional reinforcement learning problem, where the agent must learn reusable primitives that enable reaching a goal from any designated region within the environment. Training is deemed successful once the agent reaches the goal region from all ten predefined test starting positions.  
To evaluate learning efficiency, we define \textit{consumption} as the global step at which this success criterion is first achieved and the total training budget. A lower consumption indicates more efficient exploration. If the agent fails to meet the success condition within the allotted training steps, a consumption value of \textit{highest global steps} is assigned.
As shown in Fig.~\ref{fig:consumption_comparison} and Fig.~\ref{fig:efficiency}, although the agent without guidance is occasionally able to reach the goal from certain starting positions, nearly all of them fail to satisfy the overall task criterion, which is reflected in their exceedingly high step consumption.

\textbf{Success Certificate:}
We run 200 trials of the learned policies for different selected start positions and evaluate the performance with both dynamic models. The results are shown in Table~\ref{tab:success_rate}. We observe that in cluttered environments, assuming that all of the planned path is feasible and well tracked, the success rates of all the other rewards are 0, while our method and tree-based reward achieve 100\%; In cluttered environments that contain infeasible regions, all the baseline methods fail with success rates dropping to 0, while our approach consistently achieves near 100\% success. Notably, for the methods relying on RRT-generated reward, once the planned path is blocked by obstacles, the reward structure effectively degrades into a sparse, binary signal, making learning extremely difficult in such settings.

\begin{figure}[H]
\begin{center}
    \begin{minipage}{0.38\textwidth}
        \centering
        \includegraphics[width=\textwidth]{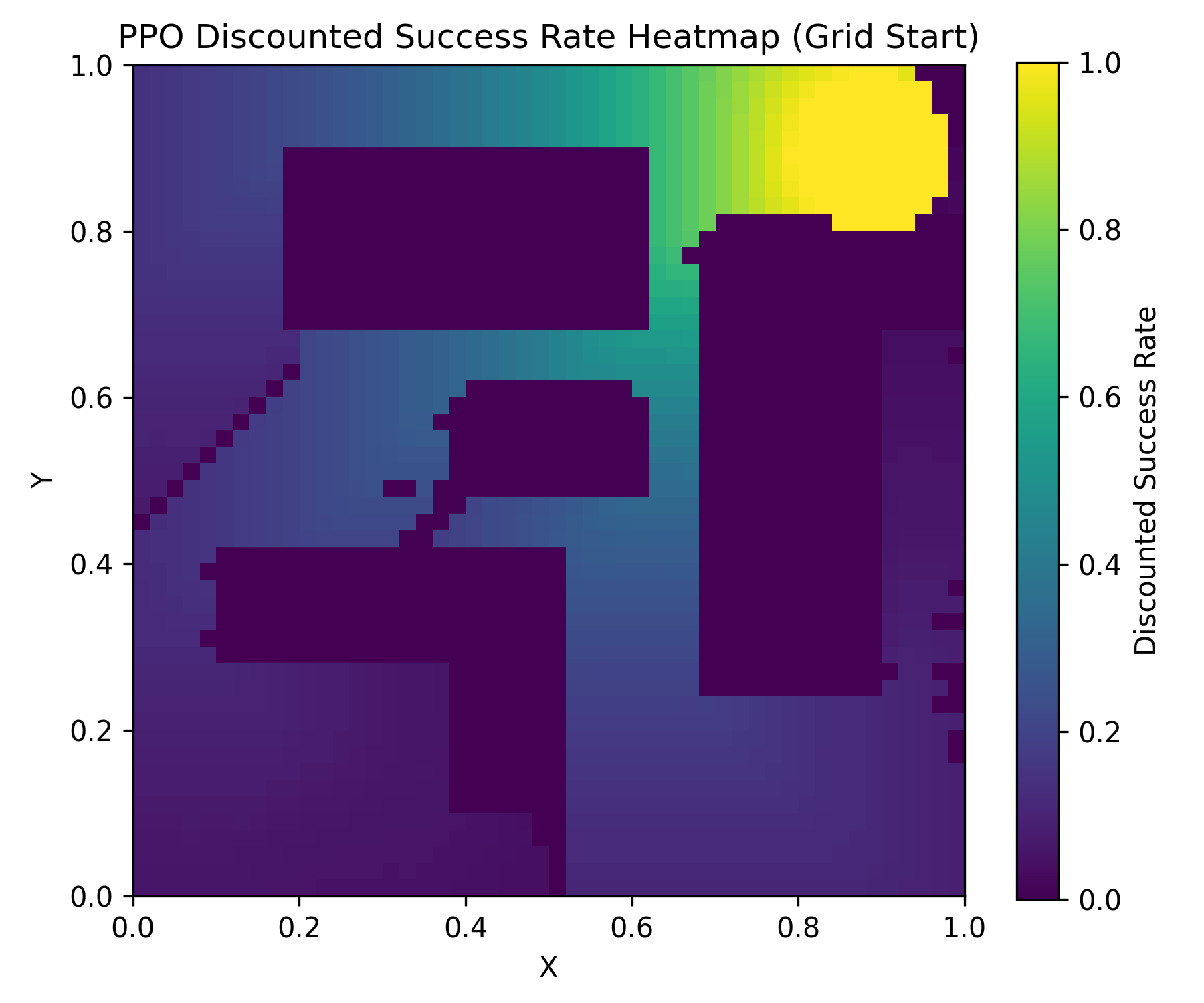}
    \end{minipage}
\begin{minipage}{0.291\textwidth}
        \centering
        \includegraphics[width=\textwidth]{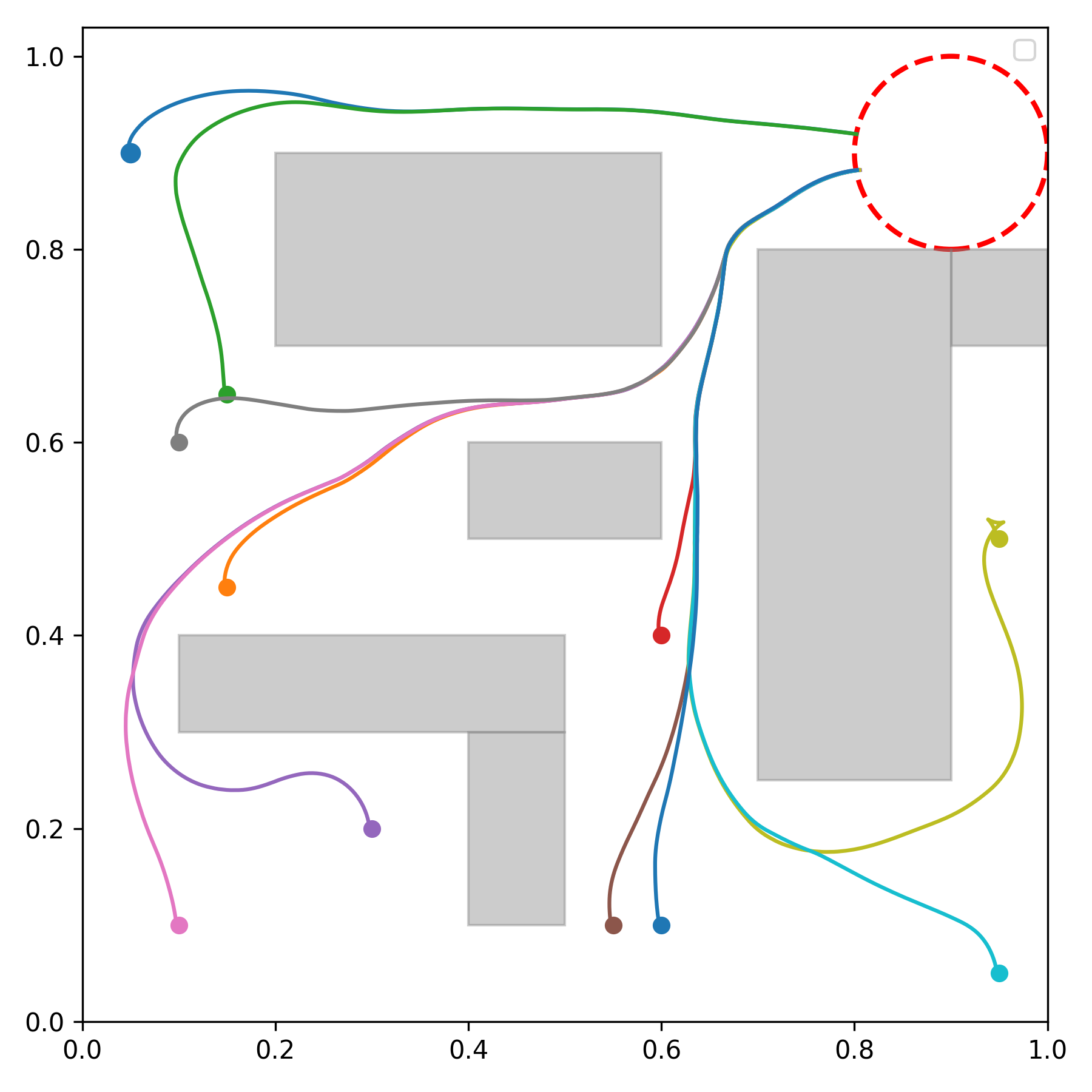}
    \end{minipage}
    \hfill
    \begin{minipage}{0.291\textwidth}
        \centering
        \includegraphics[width=\textwidth]{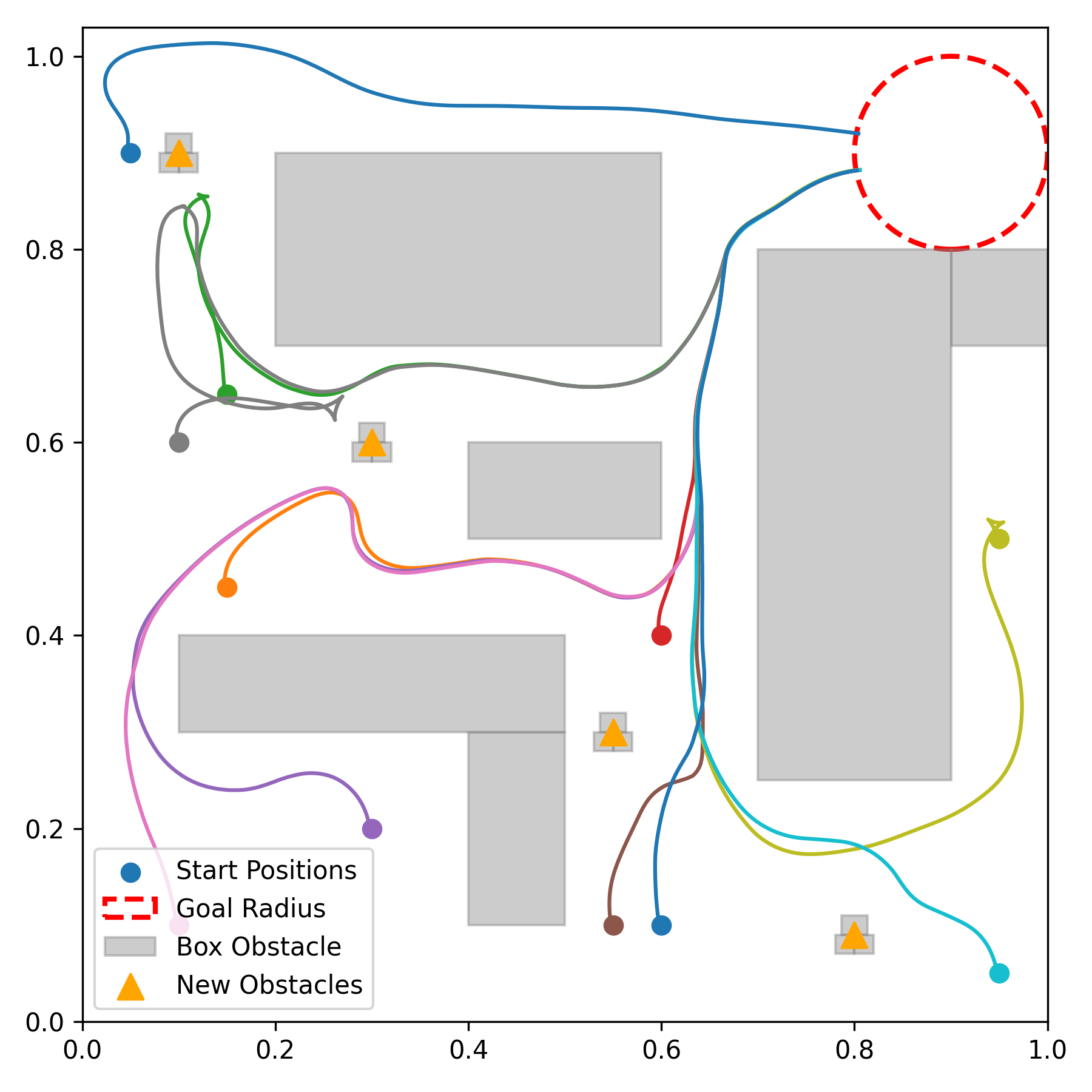}
    \end{minipage}

\caption{
Demonstration of generalization and recovery. 
\textbf{Left:} Q-map showing learned goal proximity across the state space. 
\textbf{Middle:} Trajectories from 10 random start positions reaching the goal in an undisturbed environment. 
\textbf{Right:} Recovery behavior with unseen obstacles and a distance-based safety filter.
}
\label{fig:filter}
\end{center}
\vspace{-1em}
\end{figure}

\begin{wraptable}{r}{0.40\textwidth}  
\vspace{-1em}  
\centering
\caption{Success analysis under different path feasibility settings.}

\begin{subtable}[t]{\linewidth}
\centering
\caption{Feasible-only paths}
\begin{tabular}{l|cc}
\toprule
\textbf{Method} & \textbf{Quadrotor} & \textbf{Vehicle} \\
\midrule
RRG     & \cmark & \cmark \\
RRT     & \cmark & \cmark \\
Others  & \xmark & \xmark \\
\bottomrule
\end{tabular}
\end{subtable}

\vspace{0.5em}

\begin{subtable}[t]{\linewidth}
\centering
\caption{Paths containing infeasible regions}
\begin{tabular}{l|cc}
\toprule
\textbf{Method} & \textbf{Quadrotor} & \textbf{Vehicle} \\
\midrule
RRG     & \cmark & \cmark \\
RRT     & \xmark & \xmark \\
Others  & \xmark & \xmark \\
\bottomrule
\end{tabular}
\end{subtable}
\label{tab:success_rate}
\vspace{-1em}
\end{wraptable}
\textbf{Generalization to Unseen Situations:}
As discussed in Sec.~\ref{sec:generalize}, a key advantage of our framework lies in its ability to generalize to arbitrary start states and integrate recovery mechanisms seamlessly. To evaluate this capability, we conduct a visualization experiment as shown in Fig.~\ref{fig:filter}. 
The left panel displays the discounted success score, defined as $\mathbb{I}_{\text{success}} \cdot (\text{steps})^{\gamma}$, where $\mathbb{I}_{\text{success}}$ is the indicator function for goal-reaching, ``steps'' denotes the episode length and $\gamma \in (0,1)$ is a discount factor that emphasizes earlier success.
This illustrates that the agent learns a globally coherent value landscape that supports goal-reaching from diverse initializations. In the middle panel, we visualize 10 representative trajectories from randomly selected start points, all successfully reaching the goal without any external intervention. 
To simulate realistic failures or unexpected conditions, we introduce unseen obstacles in the test environment and activate a simple distance-based safety filter to avoid entering unsafe zones. The resulting trajectories, shown in the right panel, demonstrate that the agent can dynamically adapt its path and still reach the goal, confirming the effectiveness of our method in handling perturbed and previously unseen situations. While not the primary focus, this illustrates the potential of our framework for supporting broader safety-critical extensions.

\section{Conclusion\label{sec:conclusion}}
DRL suffers from inefficient exploration, especially in cluttered environments. Compared to random exploration, motion-planning-based methods can provide effective guidance for learning. However, tree-based planners that yield a single path may fail to adapt when environmental changes, limiting their guidance. This paper proposes a novel graph-based reward scheme that maintains a model-free setup while offering global guidance coverage to encourage optimal path exploration, and significantly reduce the number of controllers required. Our approach fully leverages the advantages of model-free RL and guarantees the goal-reaching performance for the original task. 

\clearpage
\section*{Limitations and Future Work}

Our method relies on a precomputed roadmap (e.g., RRG) to provide topological guidance. Although the roadmap does not need to be globally optimal, its connectivity can influence path diversity and task coverage, particularly in extremely sparse or disconnected environments. In practice, we assume that the environment is connected and continue sampling waypoints until sufficient connectivity is achieved. Although this heuristic may not guarantee formal completeness, we find that it works reliably in all our tested environments and does not significantly impact the learning or success rate in practice.

Another experiment limitation lies in the demonstration of simple recovery mechanisms, such as distance-based filters, which may not capture more nuanced safety constraints in highly dynamic or safety-critical domains. Nonetheless, our framework is compatible with more advanced model-free or model-based recovery modules, which we leave as promising future extensions.

While our method generalizes prior work from fixed start-goal pairs to arbitrary start positions, fully randomized goal configurations remain an open challenge. For each new goal, the guidance graph must be re-searched or re-planned, and storing separate graphs for all goals is impractical. We are investigating scalable alternatives and will pursue this in future work. Despite these limitations, we find that our method demonstrates strong progress in learning temporal-logic-specified primitives, and view LTL-guided generalization as a promising direction for future work.

We acknowledge the absence of real-world experiments in this work but our method is designed with practicality in mind and does not rely on simulation-specific assumptions. We need to emphasize that the proposed framework is an exploration enhancement strategy that is fundamentally model-agnostic. It can be naturally integrated into a wide range of RL algorithms, regardless of the underlying dynamics. In future work, we plan to apply this framework to more task-specific scenarios and include real-world deployment as part of our experimental agenda. Additionally, we do not compare with model-based approaches, as our focus is on model-free settings where environment dynamics are unknown or difficult to estimate.

\bibliography{reference}

\newpage
\appendix
\section*{Appendix}
\section{Cost-to-Go Calculation for Planning-Based Rewards}\label{append:cost2goal}

Before discussing the algorithms in detail, we introduce the following algorithmic primitives used in algorithms below:

\textbf{Discretization} The function \texttt{DiscretizeEnv} divides the continuous environment $\mathcal{X}$ into a uniform grid. Each cell is classified as either \emph{free} or \emph{occupied}, depending on whether it overlaps with an obstacle. The resulting map structure serves as the input graph for A* search.

\textbf{Neighbors} Given a state $s$ and a map structure \texttt{Map}, the function \texttt{Neighbors}$(s, \text{Map})$ returns the set of neighboring states that are directly reachable from $s$. In a 4-connected grid, these are the adjacent horizontal and vertical cells; in an 8-connected grid, diagonal neighbors are also included. In a graph like RRG, neighbors are determined by edges in $E$.

\textbf{Collision check} The function \texttt{isObstacle}$(s)$ returns \texttt{true} if the state $s$ is within an obstacle region, and \texttt{false} otherwise. This function is used to prune invalid neighbors from expansion.

\textbf{Cost} The function \texttt{Cost}$(s, s')$ returns the traversal cost between two adjacent states $s$ and $s'$. This is typically set to the Euclidean or Manhattan distance between the states for grid maps, or to edge weights in roadmap graphs.

\textbf{Heuristic} The function $h(s)$ provides an admissible estimate of the cost from state $s$ to the goal state $s_{\text{goal}}$. Common choices include Manhattan or Euclidean distances in grid settings.

\textbf{Path reconstruction} Once the goal state is reached, the function \texttt{ReconstructPath}$(s_{\text{goal}})$ traces backward through the parent pointers to construct the shortest path from $s_{\text{start}}$ to $s_{\text{goal}}$. The result is a trajectory $\mathbf{X}^* = \{s_0, s_1, \dots, s_d\}$ such that $s_0 = s_{\text{start}}$ and $s_d = s_{\text{goal}}$.

\begin{algorithm}[H]
\caption{RRG with Dijkstra Distance Calculation}
\label{alg:rrgastar}
\begin{algorithmic}[1]
\State Initialize $G = (V, E)$ with start node $s_{\text{start}}$
\For{$i = 1$ to $N$}
    \State $s_{\text{rand}} \gets \text{Sample()}$
    \State $s_{\text{nearest}} \gets \text{Nearest}(V, s_{\text{rand}})$
    \State $s_{\text{new}} \gets \text{Steer}(s_{\text{nearest}}, s_{\text{rand}})$
    \If{$\text{CollisionFree}(s_{\text{nearest}}, s_{\text{new}})$}
        \State $V \gets V \cup \{s_{\text{new}}\}$
        \State $E \gets E \cup \{(s_{\text{nearest}}, s_{\text{new}})\}$
        \State $S_{\text{near}} \gets \text{Near}(V, s_{\text{new}})$
        \ForAll{$s_{\text{near}} \in S_{\text{near}}$}
            \If{$\text{CollisionFree}(s_{\text{near}}, s_{\text{new}})$}
                \State $E \gets E \cup \{(s_{\text{near}}, s_{\text{new}})\}$
            \EndIf
            \If{$\text{CollisionFree}(s_{\text{new}}, s_{\text{near}})$}
                \State $E \gets E \cup \{(s_{\text{new}}, s_{\text{near}})\}$
            \EndIf
        \EndFor
    \EndIf
\EndFor
\ForAll{$s \in V$}
    \State $d(s) \gets \text{Dijkstra}(G, s, s_{\text{goal}})$ \Comment{obtain the distance to the goal}
\EndFor
\end{algorithmic}
\end{algorithm}

As shown in Algorithm~\ref{alg:rrgastar}, the RRG algorithm begins by initializing the graph $G = (V, E)$ with the start node $s_0$ and no edges (line~1). In each iteration, a sample $s_{\text{rand}}$ is drawn, and its nearest neighbor $s_{\text{nearest}}$ is found (lines~3--4). A new state $s_{\text{new}}$ is generated by steering toward the sample (line~5), and if the path is valid, it is added to the graph along with edges to nearby nodes (lines~6--14). Once the graph is constructed, Dijkstra's algorithm (as shown in Algorithm~\ref{alg:dijkstra}) is executed to compute the cost-to-go values from each node to the goal $s_{\text{goal}}$ (lines~20--21).

\begin{algorithm}
\caption{Dijkstra's Algorithm for Cost-to-Go Computation}
\label{alg:dijkstra}
\begin{algorithmic}[1]
\Function{Dijkstra}{$G, s_{\text{goal}}$}
    \ForAll{$s \in V$}
        \State $d(s) \gets \infty$
    \EndFor
    \State $d(s_{\text{goal}}) \gets 0$
    \State Initialize priority queue $Q \gets \{s_{\text{goal}}\}$
    \While{$Q$ is not empty}
        \State $s_{\text{current}} \gets \text{ExtractMin}(Q)$
        \ForAll{$s_{\text{neighbor}}$ in Neighbors($s_{\text{current}}$)}
            \State $c \gets \text{Cost}(s_{\text{current}}, s_{\text{neighbor}})$
            \If{$d(s_{\text{neighbor}}) > d(s_{\text{current}}) + c$}
                \State $d(s_{\text{neighbor}}) \gets d(s_{\text{current}}) + c$
                \State Update $Q$ with $d(s_{\text{neighbor}})$
            \EndIf
        \EndFor
    \EndWhile
    \State \Return $d(\cdot)$
\EndFunction
\end{algorithmic}
\end{algorithm}

In addition to sampling-based planning methods such as PRM and RRG described in the main text, we also consider a sample-free alternative using a discrete grid map. In this setting, the environment is discretized into a uniform grid, and A* search is directly applied to compute feasible waypoint sequences. This approach avoids the use of sampling and graph construction, and instead leverages the regular connectivity of the grid. The complete procedure is shown in Algorithm~\ref{alg:astarsearch}

Using the map either generated by sampling based method like RRG, or use the discrete map mentioned above, Algorithm~\ref{alg:astarsearch} performs a standard A* search to compute the shortest path from a start state $s_{\text{start}}$ to a goal state $s_{\text{goal}}$. The algorithm initializes the open set with $s_{\text{start}}$ and assigns its cost-to-come $g(s_{\text{start}}) = 0$ and heuristic value $f(s_{\text{start}})$ (lines~2--4). At each iteration, the node with the smallest $f$ value is selected for expansion (line~5). If the goal is reached, the optimal path is reconstructed and returned (lines~7--8). Otherwise, the algorithm expands the current node $s_{\text{current}}$ by iterating over its neighbors as defined by the map (line~10). Invalid neighbors, such as those in obstacles or already explored, are skipped (line~13). For valid neighbors, the tentative cost is computed, and if a better path is found, the corresponding $g$, $f$, and parent values are updated (lines~16--22). This process continues until the goal is found or the open set is exhausted. If the goal is unreachable, the algorithm returns failure (line~25).

\begin{algorithm}
\caption{A* Search on General Map}
\label{alg:astarsearch}
\begin{algorithmic}[1]
\Function{A*}{$\text{Map}, s_{\text{start}}, s_{\text{goal}}$}
    \State OpenSet $\gets \{s_{\text{start}}\}$
    \State $g(s_{\text{start}}) \gets 0$
    \State $f(s_{\text{start}}) \gets g(s_{\text{start}}) + h(s_{\text{start}})$
    \While{OpenSet is not empty}
        \State $s_{\text{current}} \gets \text{argmin}_{s \in \text{OpenSet}} f(s)$
        \If{$s_{\text{current}} = s_{\text{goal}}$}
            \State \Return \text{ReconstructPath}$(s_{\text{goal}})$
        \EndIf
        \State Remove $s_{\text{current}}$ from OpenSet
        \State Add $s_{\text{current}}$ to ClosedSet
        \ForAll{$s_{\text{neighbor}}$ in Neighbors($s_{\text{current}}, \text{Map})$}
            \If{$s_{\text{neighbor}} \in$ ClosedSet or isObstacle}
                \State \textbf{continue}
            \EndIf
            \State $g_{\text{tent}} \gets g(s_{\text{current}}) + \text{Cost}(s_{\text{current}}, s_{\text{neighbor}})$
            \If{$s_{\text{neighbor}}$ not in OpenSet or $g_{\text{tent}} < g(s_{\text{neighbor}})$}
                \State $g(s_{\text{neighbor}}) \gets g_{\text{tent}}$
                \State $f(s_{\text{neighbor}}) \gets g(s_{\text{neighbor}}) + h(s_{\text{neighbor}})$
                \State \text{Parent}$(s_{\text{neighbor}}) \gets s_{\text{current}}$
                \State Add $s_{\text{neighbor}}$ to OpenSet
            \EndIf
        \EndFor
    \EndWhile
    \State \Return $\infty$ \Comment{\textbf{failure}}
\EndFunction
\end{algorithmic}
\end{algorithm}

\section{Theorem~\ref{th:markovian} proof\label{append:markovianproof}}

\begin{proof}

We first prove this reward is non-Markovian in the original state space $\setS$. With the definition of MDP~\citep{sutton2018reinforcement}, we prove it by a contradiction.

Let the full history up to $t-1$ be
$h_{t-1} := \{s_0,a_0,\dots,s_{t-1},a_{t-1}\}$.
Assume, for the sake of contradiction, that the reward function in the original state space is Markovian. That is, at each timestep $t$
\begin{equation}
    r(s_t,a_t, h_{t-1}) = r(s_t,a_t)
\end{equation}
It indicates the Markovian reward is only related to the current state and action.
Suppose the agent has already achieved one r-ball, thereby updating its internal record of $D_{min}$ (which keeps track of exploration progress). Then the agent moves back and achieves this r-ball again. For these two visits, the reward is obviously different since in the second time, the agent visits the same r-ball with the same $D_{min}$, which would not lead to a reward. 
In other words, there exist two different histories \( h_1 \) and \( h_2 \) leading to the same state-action pair \( (s, a) \) but resulting in different values of previous \( D_{min} \), causing \( r(s,a,h_1) \neq r(s,a,h_2) \). This contradicts the Markov property, proving that the reward in the original state space is inherently non-Markovian.

To address the above contradiction, we incorporate $D_{min}$ into the agent's observable state. Specifically, we extend the original state space $\setS$ to
\[
   \setS^{\times} \;=\; 
   \bigl\{ (\,s,\, D_{min}\!) \;\mid\; 
            s \in \setS,\ 
            D_{min} \in \mathbb{R}_{\ge 0}
      \bigr\}.
\]
In other words, each state in $\setS^{\times}$ explicitly tracks both the physical state $s$ \emph{and} the most up-to-date information $D_{min}$ needed for the reward function. 
In this augmented space, \emph{all} historical information necessary for computing the motion-planning-guided reward is encapsulated in the tuple $(s, D_{min})$. As a result, whenever the agent revisits a physical location, the question of whether a specific $r$-ball has been ``newly achieved'' or``already achieved'' is resolved by the corresponding $D_{min}$ values stored in the augmented state. Consequently, the reward function $\tilde{r}$ can be written as:
$\tilde{r}\bigl((s_t, D_{min}),\, a_t\bigr)$,
or, if needed, 
$\tilde{r}\bigl((s_t, D_{min}),\, a_t,\,(s_{t+1}, D_{min}')\bigr)$,
thereby depending only on the \emph{current} augmented state (and possibly the next state), making it Markovian. 

Formally, for any pair of histories $h_1$ and $h_2$ that arrive at the \emph{same} augmented state $(s_t^\times)$ and take the same action $a_t$, the updated $(s_{t+1}^\times)$ remains the same (in distribution) and so does the associated reward. This is precisely the Markov property in the augmented space:
\[
   P\bigl(\tilde{r}_t \mid s_0^\times,a_0,\dots,s_t^\times,a_t,s_{t+1}^\times\bigr) 
   \;=\;
   P\bigl(\tilde{r}_t \mid s_t^\times,a_t,s_{t+1}^\times\bigr).
\]
No additional information from the entire trajectory $h_{t-1}$ is necessary once $D_{min}$ is encoded in the state. 
\end{proof}

\section{Proposition~\ref{prop:optimalinvariance} Proof \label{append:optimalinvarianceproof}}
\begin{proof}
\begin{equation}
\hat{r}_t(s_{t})=\left\{ \begin{array}{cc}
R_{-}, & \text{if $s_t$ is in the obstacles},\\
R_{++}, & \text{if $s_t$ is in the goal area }
\\
0, & \text{otherwise,}
\end{array}\right.\label{eq:original_reward_function}
\end{equation}
Let the original reward $\hat{r}$ be defined as in Eq.~\eqref{eq:original_reward_function}, and the augmented reward $\tilde{r}$ be defined as in Eq.~\eqref{eq:reward_function}. 

We first show that there exists an optimal policy that reaches the goal under the original reward $\hat{r}$. For the MDP we described in Sec.~\ref{sec:Problem}, the existence of such a policy is guaranteed by standard results in measure-theoretic Markov decision processes (see, e.g., Theorem 3.2 in \citep{hernandez1996discrete} and Theorem  8.7.2 in \citep{puterman2014markov}).

Then we adopted Theorem 1 in \citep{Ng1999RewardShaping}, which shows that the optimal policy remains invariant if the reward transformation is of the form:
\[
R'(s, a, s') = R(s, a, s') + \gamma \Phi(s') - \Phi(s)
\]
where \(\Phi: \mathcal{S} \rightarrow \mathbb{R}\) is a potential function defined over the state space \(\mathcal{S}\), and \(\gamma\) is the discount factor.
To satisfy the theorem, it suffices to construct such a potential function \(\Phi(s)\) that renders the augmented reward a potential-based transformation.  

This condition does not hold in our case if the potential function depends on historical information such as the previous minimum distance to the goal. However, the augmented state space includes the historical minimum distance \(D_{min}\), ensuring that the potential function \(\Phi(s^{\times})\) remains a well-defined function over states. Here $D_{min}(\tau_{t}) = D_{min}^{t}$.

In our framework, we could explicitly construct a potential function:
\[
\Phi(s_{t}^{\times}) = 
\begin{cases}
C(1 - e^{-k (d_0 - D_{min}^{t})}), & \text{if } D_{min}^{t} \leq d_0 \\
0, & \text{otherwise}
\end{cases}
\]
where \(d_0 = D_{min}^{t-1}\) is the minimum distance to the goal achieved so far (included in the state representation), \(C\) is a constant, and \(k\) is a sharpness parameter.

Let $C = R_+$, $\gamma = 1$ and choose large $k \gg 0$. Define the shaping term as \(F(s^{\times},a,s^{\times'}) = \gamma \Phi(s^{\times'}) - \Phi(s^{\times})\). In the case $d' < d$, which means the agent is closer to the goal, we get $F(s^{\times},a,s^{\times'}) \approx R_+$; if $d' \geq d$, which means the agent stays or gets further from the goal, then $F(s^{\times},a,s^{\times'}) \approx 0$. Then we have the reward we defined $\tilde{r}(s_{t}^{\times}) = \hat{r}(s_{t}^{\times}) + F(s^{\times},a,s^{\times'}) $, which is exactly in the form required by the theorem 1 of \citep{Ng1999RewardShaping}.
Hence, it does not alter the optimal policy of the original MDP. 

Eventually, we show that the augmented reward accelerates convergence. Temporal-Difference (TD) learning is a core component of almost all modern reinforcement learning algorithms, serving as the basis for value updates during training.
The TD error at time step $t$ is defined as:
\[
\delta_t = r(s_{t}^{\times}) + \gamma V(s_{t+1}^{\times}) - V(s_{t}^{\times}),
\]
which serves as the core learning signal for value updates.
Let the augmented reward be \( \tilde{r} = \hat{r} + \gamma \Phi(s^{\times'}) - \Phi(s^{\times}) \). The TD error under shaping becomes:
\[
\delta'_t = \hat{r}(s_{t}^{\times}) + \gamma \Phi(s_{t+1}^{\times}) - \Phi(s_{t}^{\times}) + \gamma V(s_{t+1}^{\times}) - V(s_{t}^{\times}).
\]

In sparse-reward domains $\hat r_t=0$ for most transitions, so the shaping term $F_t$ dominates $\delta'_t$.  
Because $F_t>0$ precisely when the agent reduces its distance to the goal, the TD error now carries informative progress signals at every step.  
Provided that the learning-rate schedule $\{\alpha_t\}$ satisfies the usual Robbins–Monro conditions, the larger magnitude of these early TD errors increases the expected parameter update norm and empirically accelerates convergence, a phenomenon consistent with the convergence analysis of TD learning in~\citet{sutton2018reinforcement}.
\end{proof}

\section{Experimental Setup}
\label{sec:exp_setup}

\begin{figure}[H]
  \centering
  \begin{subfigure}{0.4\linewidth}
    \includegraphics[width=\linewidth,height=\linewidth]
    {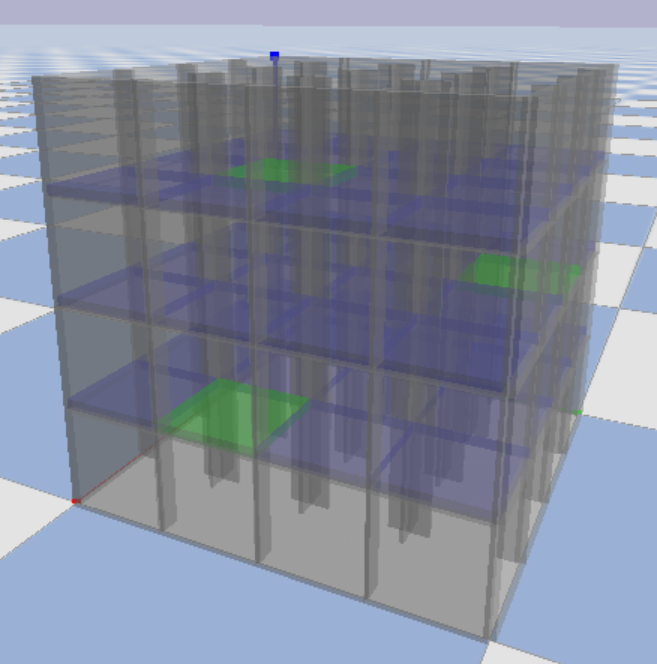}
    \caption{\textbf{Quadrotor} navigation environment.}
    \label{fig:quad_env}
  \end{subfigure}
  \hspace{0.05\linewidth}
  \begin{subfigure}{0.4\linewidth}
    \includegraphics[width=\linewidth,height=\linewidth]
    {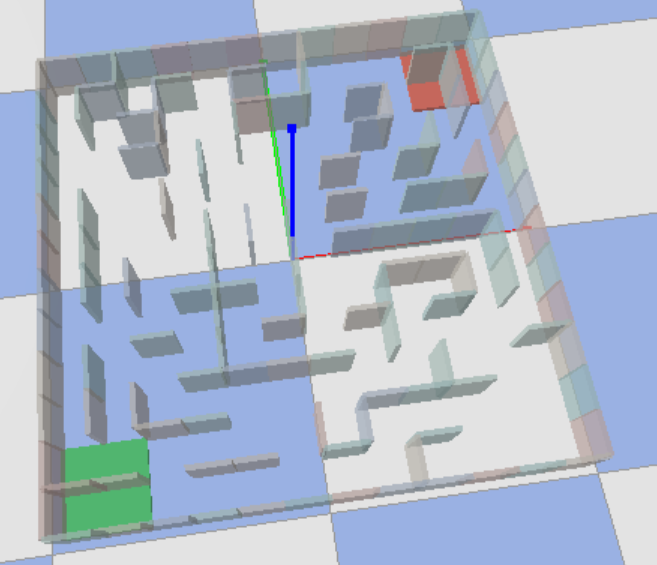}
    \caption{\textbf{Dog/Car-2D} maze environment.}
    \label{fig:dog_env}
  \end{subfigure}

  \caption{(a): a four-level $1\,\text{m}^3$ cubic maze for the \textit{Quadrotor} agent, which navigates with 3-DoF velocity commands $\langle v_x,v_y,v_z\rangle$.  
(b): a $12\times12$ planar maze with additional loops for the \textit{Dog/Car-2D} agent, controlled by forward speed and yaw rate $\langle v,\omega\rangle$.  
In both tasks the agent starts from a start region and must reach a designated goal without collisions.}
  \vspace{-1em}
  \label{fig:env_overview}
\end{figure}

We evaluate three widely-used continuous-control algorithms—PPO, SAC, and TD3-on two custom environments: a \emph{Quadrotor} navigation task (3-DoF velocity commands) and a 2-D \emph{Dog/Car} navigation task (forward velocity and yaw rate). Figure~\ref{fig:env_overview} depicts both environments.

The quadrotor operates in a maze-like workspace, which is a $1\,\mathrm{m}\times1\,\mathrm{m}\times1\,\mathrm{m}$
cube discretised into four horizontal floors, each $0.25$m high.
Every floor is a $4\times4$ room grid ($0.25$m per room).
A single \emph{vertical shaft} (highlighted in green at construction time)
links one designated room on each floor, enabling upward traversal
while all other ceiling tiles remain closed. Actions are high-level commands
$\langle v_x,v_y,v_z\rangle\in[-1.5,1.5]^3$m/s.

The ground agent navigates in a $12\times12$ perfect maze
augmented with $15\%$ additional loops (wall thickness $0.02$m,
corridor width $0.14$m).
Actions are high-level commands
$\langle v,\omega\rangle$ with $v\in[0,1.0]$m/s and
$\omega\in[-2.0,2.0]$rad/s.
Episodes start in the bottom-left green square
($0.32$m\,\texttimes\, $0.32$m) and end in the
top-right orange square of the same size.



Quadrotor agents are trained for $8{\times}10^{6}$ environment steps, while Dog/Car-2D agents are trained for $4{\times}10^{6}$ steps. All runs use four synchronous workers and log to \textsc{Weights\,\,Biases}.
Tables~\ref{tab:quad_hparams} and \ref{tab:dog_hparams} list the remaining hyper-parameters, which are shared across all baseline algorithms within each environment to ensure fair comparison.

\begin{table}[H]
\caption{Hyper-parameters for the \textbf{Quadrotor} environment.}
\label{tab:quad_hparams}
\centering
\small
\setlength{\tabcolsep}{5pt}
\renewcommand{\arraystretch}{1.05}
\begin{tabular}{lccc}
\toprule
\textbf{Hyper-parameter} & \textbf{PPO} & \textbf{SAC} & \textbf{TD3}\\
\midrule
[0.2em]
Policy network                 & $[64,\,64]$ & $[64,\,64]$ & $[64,\,64]$ \\
Value / \(Q\) network          & $[64,\,64]$ & $[64,\,64]$ & $[64,\,64]$ \\[0.2em]
Learning rate                  & $3{\times}10^{-4}$ & $3{\times}10^{-4}$ & $3{\times}10^{-4}$ \\
Batch size                     & 1024 & 1024 & 1024 \\
Replay buffer size             & — & 500\,k & 500\,k \\[0.2em]
Discount factor \(\gamma\)     & 0.99 & 0.99 & 0.99 \\
Polyak coefficient \(\tau\)    & — & 0.005 & 0.005 \\[0.2em]
Train freq / grad steps        & on-policy (256 steps) & 4 / 6 & 4 / 8 \\
Exploration term               & entropy $0.001$ & \(\alpha\) (auto) & noise \(\sigma=0.4\) \\
Total env.\ steps              & \multicolumn{3}{c}{$8{\times}10^{6}$ (4 workers)}\\
\bottomrule
\end{tabular}
\end{table}

\begin{table}[H]
\caption{Hyper-parameters for the \textbf{Dog/Car-2D} environment.}
\label{tab:dog_hparams}
\centering
\small
\setlength{\tabcolsep}{5pt}
\renewcommand{\arraystretch}{1.05}
\begin{tabular}{lccc}
\toprule
\textbf{Hyper-parameter} & \textbf{PPO} & \textbf{SAC} & \textbf{TD3}\\
\midrule
[0.2em]
Policy network                 & $[64,\,64]$ & $[64,\,64]$ & $[64,\,64]$ \\
Value / \(Q\) network          & $[64,\,64]$ & $[64,\,64]$ & $[64,\,64]$ \\[0.2em]
Learning rate                  & $3{\times}10^{-4}$ & $3{\times}10^{-4}$ & $3{\times}10^{-4}$ \\
Batch size                     & 1024 & 1024 & 1024 \\
Replay buffer size             & — & 500\,k & 500\,k \\[0.2em]
Discount factor \(\gamma\)     & 0.99 & 0.99 & 0.99 \\
Polyak coefficient \(\tau\)    & — & 0.005 & 0.005 \\[0.2em]
Train freq / grad steps        & on-policy (256 steps) & 4 / 6 & 4 / 8 \\
Exploration term               & entropy $0.01$ & \(\alpha\) (auto) & noise \(\sigma=0.4\) \\
Total env.\ steps              & \multicolumn{3}{c}{$4{\times}10^{6}$ (4 workers)}\\
\bottomrule
\end{tabular}
\end{table}

\end{document}